\documentclass[twoside,11pt]{article}

\usepackage{jmlr2e}

\makeatletter
\def\UrlAlphabet{%
      \do\a\do\b\do\c\do\d\do\e\do\f\do\g\do\h\do\i\do\j%
      \do\k\do\l\do\m\do\n\do\o\do\p\do\q\do\r\do\s\do\t%
      \do\u\do\v\do\w\do\x\do\y\do\z\do\A\do\B\do\C\do\D%
      \do\E\do\F\do\G\do\H\do\I\do\J\do\K\do\L\do\M\do\N%
      \do\O\do\P\do\Q\do\R\do\S\do\T\do\U\do\V\do\W\do\X%
      \do\Y\do\Z}
\def\UrlDigits{\do\1\do\2\do\3\do\4\do\5\do\6\do\7\do\8\do\9\do\0}
\g@addto@macro{\UrlBreaks}{\UrlOrds}
\g@addto@macro{\UrlBreaks}{\UrlAlphabet}
\g@addto@macro{\UrlBreaks}{\UrlDigits}
\makeatother
\usepackage{wrapfig}

\ShortHeadings{LibMTL: A Python Library for Multi-Task Learning}{Lin and Zhang}
\firstpageno{1}

\begin{document}

\title{LibMTL: A Python Library for Multi-Task Learning}

\author{\name Baijiong Lin$^1$ \email bj.lin.email@gmail.com \\
       \name Yu Zhang$^{1,2,}$\thanks{Corresponding author.} \email yu.zhang.ust@gmail.com \\
       \addr $^1$Department of Computer Science and Engineering,
       Southern University of Science and Technology, Shenzhen, China\\
       \addr $^2$Peng Cheng Laboratory, Shenzhen, China}

\editor{}

\maketitle

\begin{abstract}
This paper presents \texttt{LibMTL}, an open-source Python library built on \texttt{PyTorch}, which provides a unified, comprehensive, reproducible, and extensible implementation framework for Multi-Task Learning (MTL). \texttt{LibMTL} considers different settings and approaches in MTL, and it supports a large number of state-of-the-art MTL methods, including 12 loss weighting strategies, 7 architectures, and 84 combinations of different architectures and loss weighting methods. Moreover, the modular design in \texttt{LibMTL} makes it easy-to-use and well extensible, thus users can easily and fast develop new MTL methods, compare with existing MTL methods fairly, or apply MTL algorithms to real-world applications with the support of \texttt{LibMTL}. The source code and detailed documentations of \texttt{LibMTL} are available at \url{https://github.com/median-research-group/LibMTL} and \url{https://libmtl.readthedocs.io}, respectively.
\end{abstract}


\section{Introduction}

Multi-Task Learning (MTL) \citep{caruana1997multitask,ZhangY21} is an important area in both machine learning community and industrial community. By learning several related tasks simultaneously, this learning paradigm could not only improve the generalization performance but also reduce the storage cost and inference time, thus it has been applied to many real-world scenarios such as automatic driving, natural language understanding, recommendation system, robotic control, bioinformation, and so on \citep{ZhangY21}. Although many State-Of-The-Art (SOTA) MTL models have been proposed recently, most of them are implemented in their respective framework with different experimental details or there are no public implementations for them. Therefore, it is not easy to extend existing MTL algorithms to real-world applications or make a fair comparison with them when designing new MTL models.

To remedy such situation, we develop a Python library for MTL called \texttt{LibMTL}, which has three key features. Firstly, \texttt{LibMTL} provides a unified code base to cover different MTL settings such as the single-input and multi-input problems. It allows a convenient, fair, and consistent comparison between different MTL algorithms in various application scenarios. Secondly, built on PyTorch \citep{PaszkeGMLBCKLGA19}, \texttt{LibMTL} has supported lots of SOTA MTL models, especially those deep MTL models, including 12 loss weighting strategies, 7 MTL architectures, and different combinations of those two kinds of methods. Thirdly, \texttt{LibMTL} follows the modular design principles, which allows users to flexibly and conveniently add customized components or make personalized modifications. Therefore, users can easily and fast develop new MTL models or apply existing MTL algorithms to new application scenarios with the support of \texttt{LibMTL}.

\section{Settings and Approaches in MTL} \label{sec:cases}

\begin{wrapfigure}{r}{0.55\textwidth}
\vskip -0.2in
\includegraphics[width=0.55\textwidth]{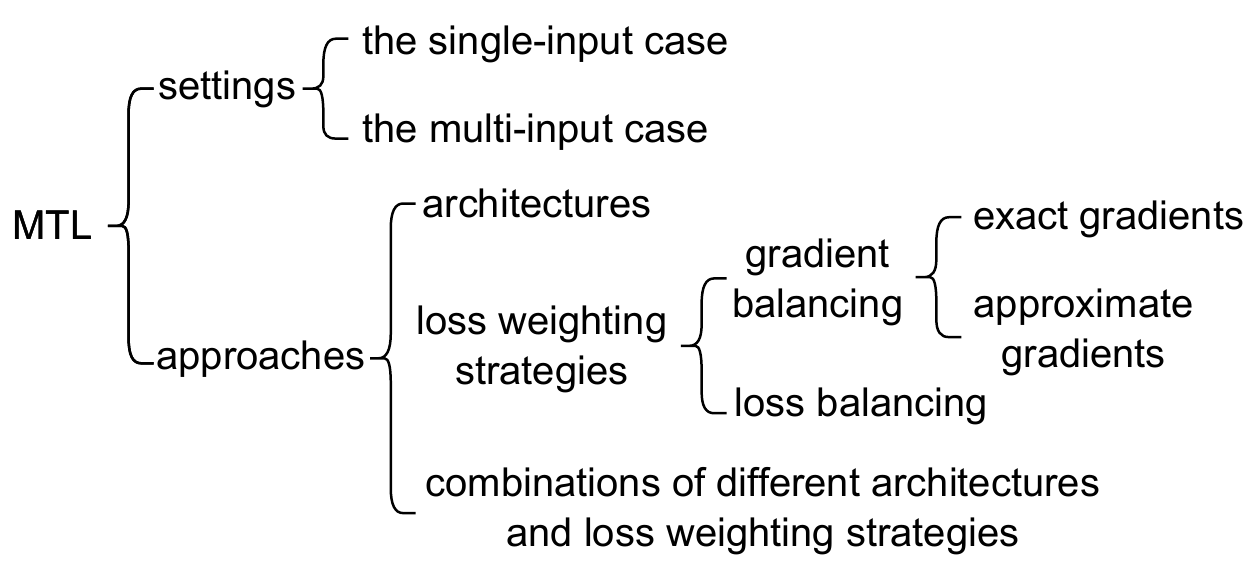}
\caption{Categories of settings and approaches in MTL.}
\label{fig:mtl}
\vskip -0.1in
\end{wrapfigure}

Suppose there are $T$ tasks and each task $t$ has its corresponding dataset $\mathcal{D}_t=\{\mathbf{X}_t, \mathbf{Y}_t\}$. Let $f(\cdot;\theta,\psi_{1:T})$ denotes an MTL model with task-shared parameters $\theta$ and task-specific parameters $\psi_{1:T}$. MTL aims to train a model $f$ on all datasets $\mathcal{D}_{1:T}$ and expects $f$ to perform well on each task. There are usually two settings in MTL: \textit{the single-input case} where each task has the same input data, i.e., $\mathbf{X}_m=\mathbf{X}_n$ for any $m\neq n$, and \textit{the multi-input case} where each task has its own input data, i.e., $\mathbf{X}_m\neq\mathbf{X}_n$ for any $m\neq n$. Those two settings reply on the application scenarios and they are different in the training implementation.

There are two main lines of research for MTL. The first line is to design the optimization strategy for MTL. Since how to balance multiple training losses in MTL directly affects the update of the task-shared parameters $\theta$, several methods are proposed to balance the losses or gradients of all the tasks in different ways, which are called loss balancing methods and gradient balancing methods, respectively. \cite{lin2021closer} have mathematically unified those two types of methods, which are different in implementations, as \textit{loss weighting strategies}. Besides, gradient balancing methods need to calculate the gradients of the task-shared parameters $\theta$ for every task, which may be computationally intensive when the number of shared parameters or tasks is large. Thus, \cite{sk18} propose to use gradients of feature representations to approximate the exact gradients of shared parameters, which significantly reduces the computational cost and is followed by other gradient balancing methods such as \citep{ChenNHLKCA20, liu2021imtl}. Obviously, those two ways to calculate gradients are different in implementations. The second line is to design the \textit{architecture} in deep neural networks for MTL and it directly determines which parameters are shared and how to share.

Noticeably, those two lines of research are almost orthogonal to each other as the loss weighting methods are mainly related to the objective function while the design of the architecture is to learn relationships between tasks. %the architectures are agnostic with the loss weighting methods because they independently affect the forward pass in the training process while the loss weighting methods only influence the backward pass.
Thus, loss weighting strategies can be seamlessly combined with architectures to further improve the performance of MTL.

To summarize, as shown in Figure \ref{fig:mtl}, MTL has two settings and its learning approaches can be divided into three categories. %there are many different use cases in MTL when using different MTL methods or applying them to different applications.

\section{The \texttt{LibMTL} Library}

In this section, we introduce the \texttt{LibMTL} library, which provides a unified and easy-to-use framework for MTL as mentioned in Section \ref{sec:cases}. In Section \ref{sec:support_methods}, we introduce MTL models implemented in \texttt{LibMTL}, which enables consistent and reproducible comparisons between different MTL models. In Section \ref{sec:modular_design}, we
present the modular design in \texttt{LibMTL}, which allows flexible and extensible customization for new MTL methods or potential MTL applications. Finally, we show that \texttt{LibMTL} is more comprehensive and up-to-date than the existing MTL libraries in Section \ref{sec:related_libraries}.

\subsection{Supported MTL Methods} \label{sec:support_methods}
Currently, \texttt{LibMTL} supports 12 loss weighting strategies, namely, Equal Weighting (\textbf{EW}), Gradient Normalization (\textbf{GradNorm}) \citep{chen2018gradnorm}, Uncertainty Weights (\textbf{UW}) \citep{kgc18}, \textbf{MGDA} \citep{sk18}, Dynamic Weight Average (\textbf{DWA}) \citep{ljd19}, Geometric Loss Strategy (\textbf{GLS}) \citep{ChennupatiSYR19}, Projecting Conflicting Gradient (\textbf{PCGrad}) \citep{pcgrad}, Gradient sign Dropout (\textbf{GradDrop}) \citep{ChenNHLKCA20}, Impartial Multi-Task Learning (\textbf{IMTL}) \citep{liu2021imtl}, Gradient Vaccine (\textbf{GradVac}) \citep{wang2021gradient}, Conflict-Averse Gradient descent (\textbf{CAGrad}) \citep{liu2021conflict}, and Random Loss Weighting (\textbf{RLW}) \citep{lin2021closer}. Moreover, it supports 7 MTL architectures, i.e., Hard Parameter Sharing (\textbf{HPS}) \citep{Caruana93}, \textbf{Cross-stitch} Networks \citep{MisraSGH16}, Multi-gate Mixture-of-Experts (\textbf{MMoE}) \citep{ma2018modeling}, Multi-Task Attention Network (\textbf{MTAN}) \citep{ljd19}, Customized Gate Control (\textbf{CGC}) \citep{tang2020progressive}, Progressive Layered Extraction (\textbf{PLE}) \citep{tang2020progressive}, \textbf{DSelect-k} \citep{hazimeh2021dselect}. Most of the aforementioned methods have no official implementation and are implemented by ourselves. Besides, \texttt{LibMTL} supports combinations of each loss weighting strategy and each architecture, leading to 84 combinations of them in total.

\subsection{The Modular Design of \texttt{LibMTL}} \label{sec:modular_design}

\begin{figure}[!htbp]
\centering
\includegraphics[width=\textwidth]{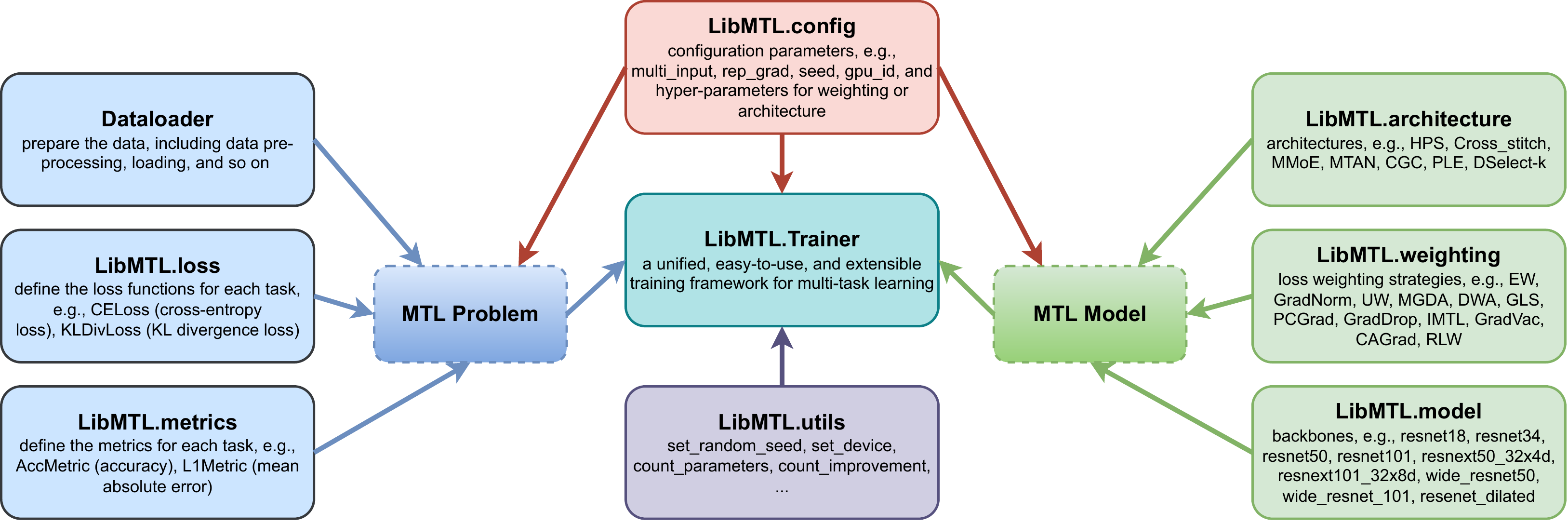}
\caption{The overall framework of \texttt{LibMTL}.}
\label{fig:framework}
\end{figure}

Figure \ref{fig:framework} shows the overall framework of \texttt{LibMTL}, which is divided into different functional modules to allow users to flexibly and conveniently add customized designs or modifications in any module.

In \texttt{LibMTL}, each module has different functionalities. The \texttt{Dataloader} module is responsible for data pre-processing and loading. The \texttt{LibMTL.loss} module defines loss functions for each task. The \texttt{LibMTL.metrics} module defines evaluation metrics for all the tasks. The above three modules are highly dependent on the MTL problem under investigation. The \texttt{LibMTL.config} module is responsible for all the configuration parameters involved in the training process, such as the corresponding MTL setting (i.e. the multi-input case or not), the potential hyper-parameters of loss weighting strategies and architectures, the training configuration (e.g., the batch size, the running epoch, the random seed, and the learning rate), and so on. This module adopts command-line arguments to enable users to conveniently set those configuration parameters. The \texttt{LibMTL.Trainer} module provides a unified framework for the training process under different MTL settings and for different MTL approaches as introduced in Section \ref{sec:cases}. The \texttt{LibMTL.utils} module implements some useful functionalities for the training process such as calculating the total number of parameters in an MTL model. The \texttt{LibMTL.architecture} and \texttt{LibMTL.weighting} modules contain the implementations of various architectures and loss weighting strategies, respectively, as introduced in Section \ref{sec:support_methods}. The \texttt{LibMTL.model} module includes some popular backbone networks (e.g., ResNet). The last three modules are highly related to MTL models.

Noticeably, such modular design makes \texttt{LibMTL} easy-to-use and well extensible. For example, when applying to new applications, users only need to prepare the new dataloaders and select (or re-define) appropriate loss and metric functions, and they can apply existing MTL methods implemented in \texttt{LibMTL}.
Besides, for researchers to develop new MTL methods such as new architectures, they can easily implement their new method with the support of \texttt{LibMTL}, make a fair comparison with existing models, and combine the new architecture with modern loss weighting methods to further improve the performance based on \texttt{LibMTL}.

\subsection{Comparison to Related Libraries} \label{sec:related_libraries}

There are some libraries that have been developed for MTL recently. For example, RMTL \citep{cao2019rmtl} is implemented in R to support shallow MTL methods such as linear regularized methods. Another library, i.e., MTLV \citep{rahimi2021mtlv}, only provides a few MTL architectures for natural language processing. Compared with them, \texttt{LibMTL} is more comprehensive and up-to-date. Firstly, \texttt{LibMTL} covers more settings and approaches as introduced in Section \ref{sec:cases}, which means that \texttt{LibMTL} can be applied to more application scenarios. Secondly, \texttt{LibMTL} implements more SOTA MTL models, especially those based on deep neural networks.

\section{Conclusion}
We present \texttt{LibMTL}, a comprehensive and extensible library for MTL. Built on \texttt{PyTorch}, it provides a unified training framework for different settings in MTL and possesses many SOTA MTL algorithms. In our future work, we will continuously maintain this library to incorporate newly proposed MTL models, update the documentations, and add more applications from different areas.

\acks{This work is supported by NSFC key grant 62136005 and NSFC general grant 62076118.}

\bibliography{LibMTL}

\end{document}